\documentclass[runningheads]{llncs}

\usepackage[T1]{fontenc}
\def\doi#1{\href{https://doi.org/\detokenize{#1}}{\url{https://doi.org/\detokenize{#1}}}}
\usepackage{graphicx}
\usepackage{listings}
\usepackage{amsfonts}
\usepackage{amsmath}
\usepackage{graphicx}
\usepackage{cite}
\usepackage{subcaption}
\usepackage[font={small}]{caption}
\usepackage{algorithm,algpseudocode}
\usepackage{siunitx}
\usepackage{makecell}

\usepackage[table,xcdraw]{xcolor}
\usepackage{flushend} 
\usepackage{multicol}
\usepackage{booktabs}
\usepackage{pifont}
\usepackage{wrapfig}

\usepackage{enumitem}
\usepackage[pagebackref=false,breaklinks=true,colorlinks,urlcolor=blue,citecolor=blue,linkcolor=blue,bookmarks=false]{hyperref}

\newcommand{\tickmark}{\ding{51}}%
\newcommand{\crossmark}{\ding{55}}%

\DeclareCaptionFormat{subfig}{\figurename~#1#2#3}
\DeclareCaptionSubType*{figure}

\newcommand{\norm}[1]{\left\lVert#1\right\rVert}
\newcommand{\algoName}{GLiDE}

\begin{document}
\title{GLiDE: Generalizable Quadrupedal Locomotion in Diverse Environments with a Centroidal Model}
\titlerunning{GLiDE}
%
\author{Zhaoming Xie\inst{1, 2} \and
Xingye Da\inst{2} \and
Buck Babich\inst{2} \and \\ Animesh Garg\inst{2, 3} \and Michiel van de Panne\inst{1}}
\authorrunning{Xie et al.}
%
\institute{University of British Columbia \and
NVIDIA \and University of Toronto, Vector Institute\\}
\maketitle              
\begin{abstract}
Model-free reinforcement learning (RL) for legged locomotion commonly relies on a physics simulator that can 
accurately predict the behaviors of every degree of freedom of the robot. 
In contrast, approximate reduced-order models are commonly used for many model predictive control strategies. 
In this work we abandon the conventional use of high-fidelity dynamics models in RL and we instead
seek to understand what can be achieved when using RL with a much simpler centroidal model when applied to quadrupedal locomotion. 
We show that RL-based control of the accelerations of a centroidal model is surprisingly effective, 
when combined with a quadratic program to realize the commanded actions via ground contact forces.
It allows for a simple reward structure, reduced computational costs, and robust sim-to-real transfer. 
We show the generality of the method by demonstrating flat-terrain gaits, stepping-stone locomotion, 
two-legged in-place balance, balance beam locomotion, and direct sim-to-real transfer. 

\end{abstract}
\section{Introduction}
Tremendous progress has been made recently in legged locomotion, as achieved using both model predictive control (MPC) and reinforcement learning (RL). MPC methods leverage modern optimization techniques and known models of the physics, possibly simplified, to synthesize responsive control at run time. However, they can be prone to local minima, can require substantial manual tuning, and are difficult to generalize to complex terrains and rich perceptual streams. Moreover, real-time MPC often uses linear models for computational efficiency, making it non-trivial to represent the legged system's nonlinear and hybrid nature. Alternatively, model-free methods such as RL utilize Monte-Carlo sampling strategies and can learn policies for general tasks. This comes at the expense of 
careful system modeling,
extensive offline simulation using the full dynamics model,  
and careful reward-crafting to produce results that are non-destructive for physical robots.

\begin{figure}[t]
    \centering
    \includegraphics[width=0.35\columnwidth]{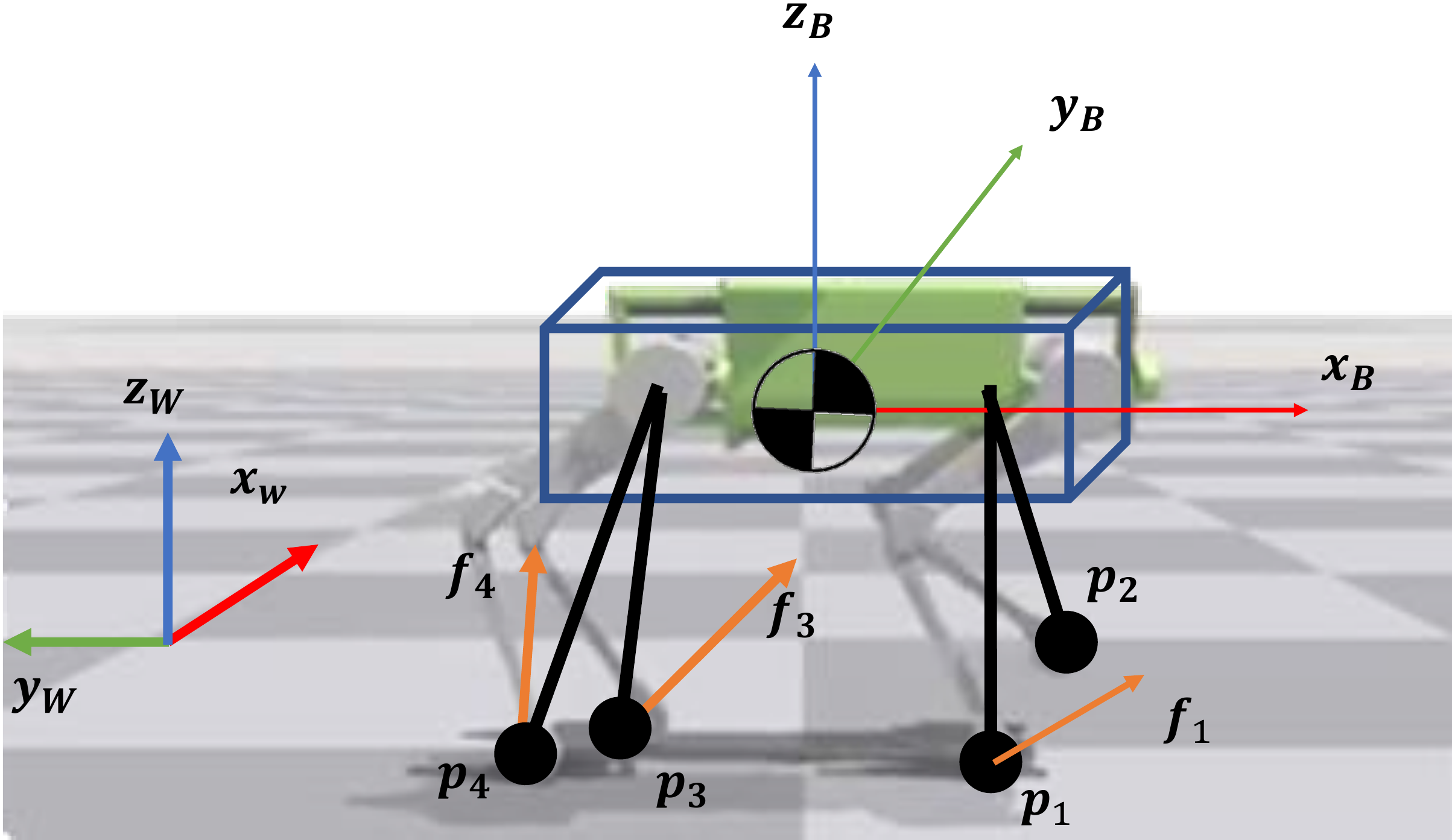}
    \caption{
    The centroidal model consists of a rigid body with virtual legs. 
    Control is realized by generating ground reaction forces at foot locations.}
    \label{fig:centroidal}
\end{figure}

\begin{figure*}
    \centering
    \includegraphics[width=0.8\textwidth]{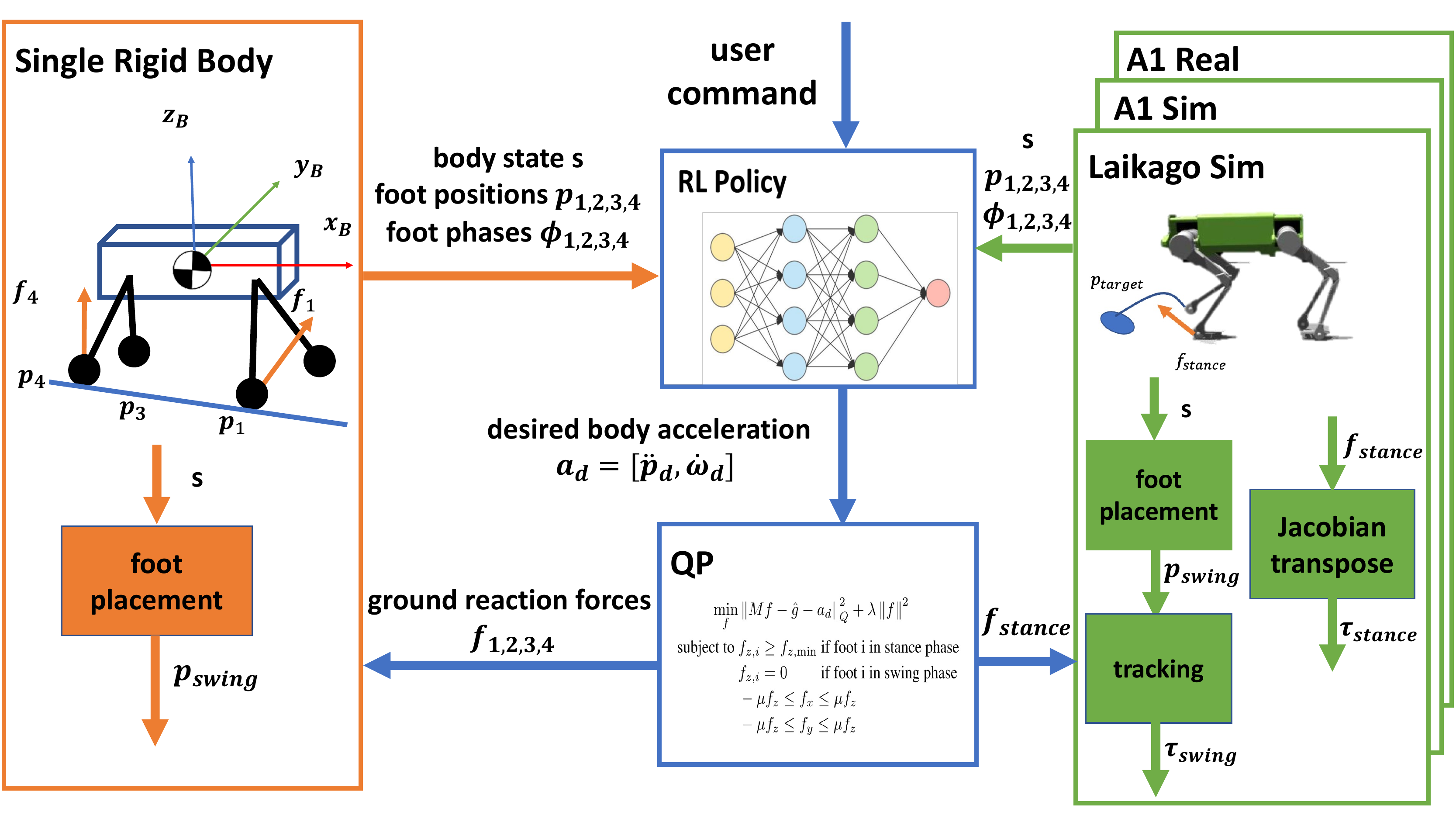}
    \caption{Overview of our system. A control policy is trained with RL on the centroidal dynamics. The policy generates desired body acceleration commands, which are transformed to ground reaction forces using a QP. The resulting policy is then realized on the full-order model using the Jacobian transpose for stance leg control and trajectory tracking for swing leg control. All components are executed at $\SI{100}{\hertz}$.}
    \label{fig:system}
\end{figure*}

In this paper, we present \algoName, Generalizable Quadrupedal Locomotion in Diverse Environments with a Centroidal Model, where we eschew the conventional wisdom of applying RL to a high-fidelity model, and
instead we ask what can be achieved when RL is used with a highly-abstracted centroidal model,
as shown in Fig.~\ref{fig:centroidal}.
The action consists of the body linear and angular accelerations,
which are realized by the ground reaction forces (GRFs), as computed via a quadratic program (QP).
We further assume a specified gait pattern and a foot-placement function.  
Taken together, these choices allow for a simple task reward specification, in contrast to the more complex reward commonly required for full-model RL. Additional constraints such as no-slip constraints and leg lengths are enforced via the QP and the foot placement.  
The GRFs are realized on the full model using the Jacobian transpose for the stance legs 
to compute joint torques, and a trajectory-tracking approach for the swing legs. A system overview is given in Fig.~\ref{fig:system}. The resulting policies are validated on simulations of the Laikago and A1\footnote{Laikago and A1 are quadrupedal robots made by Unitree Robotics.} robots as well as on a physical A1 robot. Please refer to the video for these results.



Our core contributions are as follows:
\begin{itemize}[leftmargin=0.5cm]
 \item We introduce an RL framework for learning control policies suited for centroidal dynamics models.
 This enables the anticipatory behavior required for quadrupedal locomotion and balance tasks, and 
 is therefore a promising alternative to RL methods that require accurate full body dynamics models.  
 The benefits of using a centroidal model include simple reward design, efficient simulation during training, and flexible and robust motion control.
    \item We introduce a novel action space for quadrupedal RL that operates in the centroidal model of the robot. Conversion between centroidal action space and the joint action space is realized via control methods such as quadratic programming and Jacobian transpose.
    \item We demonstrate the effectiveness of our approach by solving locomotion with multiple gaits, stepping stone scenarios, balance beam locomotion, and two-legged balancing.
    We also show successful transfer to a physical robot.
\end{itemize}

\section{Related Work}

Our work combines techniques used in quadrupedal locomotion in MPC and learning literature to solve challenging tasks. A comparison with some key papers is provided in Table~\ref{tab:related_work_comparison}. In this section we review more related work in this area.

\begin{table}[t]
\tiny
\begin{center}
\begin{tabular}{l|c|c|c|c}
\toprule
\rowcolor[HTML]{CBCEFB}
& \textbf{Model Used} &\makecell{\textbf{Reward} \\\textbf{/Cost Function}} & \textbf{Demonstrated Tasks} & \makecell{\textbf{Training Time}\\ \textbf{(Flat Terrain)}}\\
 \midrule
 \begin{tabular}[c]{@{}l@{}}Centroidal MPC \\ \cite{convex-mpc, representation-free-mpc, quadruped_centroidal}\end{tabular}
  & \multicolumn{1}{c|}{\begin{tabular}[c]{@{}c@{}}Linearized \\Centroidal ({\color{teal} \tickmark}) \end{tabular}} & 
  \multicolumn{1}{c|}{\begin{tabular}[c]{@{}c@{}}Linear Quadratic \\Regulator ({\color{teal} \tickmark}) \end{tabular}}
  & Flat, Back Flip & n/a\\
\rowcolor[HTML]{EFEFEF} 

 \begin{tabular}[c]{@{}l@{}}Centroidal PD \\ \cite{cheetah3_design_control, contact-adaptive, StarlEth}\end{tabular}
  & Centroidal & 
  n/a
  & Flat & n/a\\
\begin{tabular}[c]{@{}l@{}}Laikago RL \\ \cite{2020-RSS-laikago, 2021-icra-dynamics_randomization}\end{tabular} & Full Physics & 12 terms ({\color{red}\crossmark}) & Flat  & 6-8 hours\\
\rowcolor[HTML]{EFEFEF} 
\begin{tabular}[c]{@{}l@{}}Anymal RL \\ \cite{2020-icra-constraintQuadruped, 2019-science-sim2realAnyaml, 2020-science-blindQuadruped}\end{tabular}
 & Full Physics& 9-12 terms ({\color{red} \crossmark})& Flat, Get Up, Rough Terrain & 3-4 hours\\
\begin{tabular}[c]{@{}l@{}}DeepGait \\ \cite{deepgait}\end{tabular}
 & \multicolumn{1}{c|}{\begin{tabular}[c]{@{}c@{}}Centroidal \\+ Full Physics ({\color{teal} \tickmark}) \end{tabular}}& 10 terms ({\color{red}\crossmark}) & Flat, Stepping Stone, Stairs & 58 hours\\
\rowcolor[HTML]{EFEFEF} 
\begin{tabular}[c]{@{}l@{}}\textbf{\algoName} \\ \textbf{(ours)}\end{tabular} & Centroidal & 6 terms ({\color{teal}\checkmark}) & 
 \multicolumn{1}{c|}{\begin{tabular}[c]{@{}c@{}}Flat, Two-Legged Balancing \\ Stepping Stone, Balance Beam \end{tabular}}
 & 1-2 hours\\
\bottomrule
\end{tabular}
\caption{We compare with recent work on quadrupedal locomotion. MPC often uses a centroidal model for efficient control synthesis but frequently fails to generalize to challenging scenarios. RL makes use of accurate full-body simulations and can handle challenging scenarios at the expense of significant computational cost and complicated reward structures. We use the centroidal model for RL to efficiently train policies for challenging tasks with a simple reward function.}
\label{tab:related_work_comparison}
\end{center}
\end{table}

\paragraph{Reduced-Order Models for Legged Locomotion}
Due to the complexity of legged robots, MPC approaches often make use of reduced-order models for control synthesis. 
An inverted pendulum and variations thereof are commonly used as abstract models of bipedal robots to synthesize control policies, e.g., \cite{LIP-Cassie, SLIP-Cassie, LIP,  capture-point, H-LIP-Cassie}. A centroidal model can be used to generate control policies for bipeds~\cite{atlas_centroidal, atlas_centroidal2, atlas_centroidal3, centroidal_biped} and quadrupeds~\cite{convex-mpc,  representation-free-mpc, StarlEth, quadruped_centroidal}. To further simplify control synthesis, reference motions based on these models are generated offline, and feedback is then realized around these assuming local linearity. 
Direct nonlinear optimization can also be used for MPC at the cost of a slower control frequency~\cite{frequency-MPC} or additional heuristic objectives to guide the optimization~\cite{heuristic-MPC} out of local minima. Although centroidal model has been used extensively in the MPC literature, it is rarely used for RL, where a very detailed, accurate model is often assumed to be necessary. In this paper, we employ the centroidal dynamics model for RL. The resulting policy directly optimizes for long-horizon nonlinear problems without the aid of a reference trajectory or heuristic cost, and can traverse more challenging terrains such as stepping stone compared to MPC approaches.

\vspace{-5pt}

\paragraph{Deep Reinforcement Learning for Quadrupedal Robots}
Deep RL has become a viable approach for synthesizing control policies for quadrupedal robots. It is often realized via a comprehensive simulation of the robots, e.g., \cite{2019-science-sim2realAnyaml, linear-slope, 2018-rss-sim2realquadruped}. Rewards designed based on reference trajectories~\cite{2020-RSS-laikago, 2021-icra-dynamics_randomization} or carefully tuned reward terms \cite{2020-icra-constraintQuadruped, 2020-science-blindQuadruped, multiexpert-quadruped,bharadhwaj2021csc} are necessary to regularize undesirable behaviors as to be feasible for a physical robot. Training a policy often requires millions to billions of transition tuples of the full physics simulation. In this paper, we do not rely on a multibody dynamics simulator to train our policies. Instead, we simulate a low-cost centroidal dynamics model, which can be easily parallelized without resorting to special-purpose GPU physics simulators, e.g., \cite{neuralsim, 2020-IsaacGym}.  When used in combination with a simple foot placement strategy, it requires significantly less computation for scenarios that require complex contact modeling, such as stepping stone and balance beam traversal, as compared to recent work, e.g., \cite{deepgait, pixels-to-legs}.

Our work has some parallels with model-based RL, as we do not rely on a full physics simulator or a physical robot to perform rollouts. Recent work using model-based RL for legged robot control still needs to collect simulation data or physical robot data to learn either a transition model that contains the full state~\cite{model-based-legged} or center of mass ~\cite{planning-latent-action} information. We rely instead on the simple centroidal dynamics to perform rollouts, which does not require data from a full model or a physical robot. Our simulation is easily adapted to robots with different physical parameters or morphology by changing the compact set of parameters that define the centroidal model, such as the mass, inertia.

\vspace{-5pt}
\paragraph{Combination of MPC and Learning}
MPC comes with the benefit of explainability but often at the expense of significant online computation, 
manual tuning, and feature engineering. Alternatively, learned RL policies enable fast online computation and can in principal cope with complex terrains and rich perceptual observations but suffer from expensive offline computation and a control policy that is highly specific to the given robot morphology. 

A combination of MPC and RL can offer the best of both approaches. 
The use of RL in task spaces together with an MPC controller has yielded generalization in learning for manipulation skills~\cite{martin2019variable, martin2021laser}.
A MPC controller can be used to collect data offline and then efficiently queried online using supervised learning, e.g., \cite{MPC-net,supervised-learning-Marlo, centroidal-planner}. Heuristics or dynamics can be learned to enable faster optimization or planning, e.g., \cite{heuristic-MPC, learned-centroidal-dynamics-prediction}. Hierarchical control structures have been proposed to allow MPC controllers and RL policies to operate at different time scales to leverage their respective advantages, e.g., \cite{contact-adaptive, RL-atrias, hierachy-manipulation, DeepQ-Stepper, cassie_task_space}. 

Our work also combines RL and MPC. A high-level learned policy operates in the space of centroidal dynamics, producing desired linear and angular accelerations. The low-level control policy uses MPC based methods to realize these high-level commands, aided
by a quadratic program that transcribes the desired accelerations into appropriate ground reaction forces. 


\section{\algoName: RL with Centroidal Model}


In this section, we describe \algoName~for learning policies using a centroidal dynamics model and the related approach for realizing the resulting policy on the full robot model. We describe the centroidal dynamics simulation, the control of the centroidal dynamics by transforming the desired body acceleration to ground reaction forces via a quadratic program (QP), and the use of a Raibert-style heuristic for foot placement. Further, we detail the procedure for training RL policies using \algoName, and policy transfer to full-order model using MPC based approaches.
The system is outlined in Fig.~\ref{fig:system}.

\vspace{-5pt}
\paragraph{Centroidal Dynamics Model With Virtual Legs}
Our centroidal dynamics model consists of a single rigid body with four massless legs attached; see Fig.~\ref{fig:centroidal}. Note that this is a fixed centroidal model, where we ignore the dependency of the COM position and inertia tensor on the pose of the legs, as a result of massless legs assumption. Related literature also refers to this as a single rigid body model. The rigid body has mass $m$ and inertia $I \in \mathbb R^{3 \times 3}$. 
Each leg has a phase variable $\phi$ that advances linearly in time. The state of the rigid body $s = [p, \dot{p}, R, \omega]$ consists of the linear position and velocity $p, \dot{p} \in \mathbb R^3$, orientation $R \in SO(3)$, and angular velocity $\omega \in \mathbb R^3$. Following \cite{representation-free-mpc}, we represent $R$ with a rotation matrix.

We follow the derivation from \cite{representation-free-mpc} to simulate the centroidal dynamics. For completeness, we describe our notation and key equations used for simulation. For each leg $i$, ground reaction forces $f_i \in \mathbb R^3$ can be generated to drive the state of the rigid body. When walking on flat ground, if leg $i$ is in swing phase, $f_i = 0$; otherwise $f_i$ needs to obey the friction cone constraints in order to prescribe feasible forces. Given the foot position $p_i \in \mathbb R^3$ and $f_i$ for each leg $i$, the net force and torque $f_\text{net}, \tau_\text{net} \in \mathbb R^3$ can be calculated as: 
\vspace{-0.15cm}
\begin{align}
\label{eq:net_force}
\begin{split}
 f_\text{net} = \sum_{i=1}^4 f_i - mg \text{,    }\tau_\text{net} = \sum_{i=1}^4 \hat{r_i}f_i
\end{split}
\end{align}
where $g$ is the gravitational constant, $r_i = p_i - p$, and $\hat{\cdot}: \mathbb R^3 \to \mathfrak{so}(3)$ defines 
a mapping from a vector to its skew-symmetric matrix form.

The Euler integration update of the rigid body state can then be written as
\begin{align}
\label{eq:euler_integration}
\begin{split}
 p &= p + \dot{p}\Delta t\\
 \dot{p} &= \dot{p} + \frac{1}{m}f_\text{net}\Delta t\\
 R &= R\exp(\widehat{\omega\Delta t})\\
  \omega &= \omega + I^{-1}(R^T\tau_\text{net}-\hat{\omega}I\omega)\Delta t
\end{split}
\end{align}
where $\Delta t$ is the simulation time step and the exponential map $\exp: \mathfrak{so}(3) \to SO(3)$ ensures that $R$ stays on the $SO(3)$ manifold. 

We define a phase $\phi_i$ for each foot $i$ to be an integer that increments by $1$ at each time step. We define a cycle time, $T$, and a swing time, $T_\text{swing}$. Foot $i$ is in swing phase if $(\phi_i\mod T) \leq T_\text{swing}$ and in stance phase otherwise.


\vspace{-5pt}

\paragraph{Quadratic Programming Based Control}
While the centroidal model is driven by the ground reaction forces (GRFs), it is often unintuitive to directly provide the set of GRFs. Furthermore, in the RL setting, it is difficult to guarantee inequality constraints such as a friction cone if we use the GRFs as the control action. 
Instead, we choose the control action to be the desired acceleration $a_d = [\ddot p_d, \dot \omega_d]$. 
We then solve a QP to transcribe this into a set of GRFs: $f = [f_1, f_2, f_3, f_4]$ that satisfy an approximation of friction cone constraints:

\vspace{-15pt}
\begin{align}
\label{eq:QP}
\begin{split}
\mathop{\mathrm{min}}_{f} & \norm{Af - \hat{g} - a_d}_Q^2+\lambda\norm{f}^2 \\
\mathop{\text{subject to  }} & f_{z,i} \geq f_{z, \min} \text{  if foot i in stance phase}\\
& f_{z,i} = 0 \quad\quad\text{  if foot i in swing phase}\\
& -\mu f_{z} \leq f_{x} \leq \mu f_{z}\\
& -\mu f_{z} \leq f_{y} \leq \mu f_{z}
\end{split}
\end{align}
where $A \in \mathbb R^{6 \times 12}$ is the matrix that transforms GRFs to body acceleration, which is a function of foot position $p_{foot}$ and the centroidal state $s$, $\hat{g}=[g, 0_3]\in \mathbb R^6$ is the augmented gravity vector, $Q \succ 0$ is for weight adjustment of the cost terms, $\mu$ is the friction coefficient of the terrain and $\lambda > 0$ regularizes the GRFs used. The resulting GRFs are then used to simulate the centroidal dynamics.

\vspace{-5pt}
\paragraph{Foot Placement Strategy}
We use a simple Raibert-style heuristic~\cite{raibert} for foot placement. 
Specifically, in the centroidal model, for each foot $i$, if $(\phi_i \mod T) = 0$, 
i.e., at the instant where it switches from stance foot to swing foot, 
the foot position relative to the body on the $xy$-plane, $r_{i,xy}$, will be updated following:
\begin{align}
\label{eq:foot_placement}
\begin{split}
    r_{i, x} = r_{\text{ref},i, x} + k_{\text{foot},x}\,\dot p_{ x}, \text{,    }
    r_{i, y} = r_{\text{ref},i, y} + k_{\text{foot},y}\,\dot p_{ y},
\end{split}
\end{align}
where $r_{\text{ref}, i}$ are the default foot placements, typically obtained from the neutral standing pose of the robot, and $k_{\text{foot}}$ is a constant that adjusts the foot placement based on the velocity of the body $\dot p_{xy}$. In addition, a leg-length constraint is applied such that when $\norm{p_\text{foot} - p} \geq l_\text{max}$ for some maximum leg length $l_\text{max}$, the point closest to $p_\text{foot}$ that satisfies the leg length constraint will be used. 
Since the GRF for the swing foot is zero, the swing foot is assumed to not affect the dynamics of the centroidal model until it becomes the stance foot again. Importantly, the control policy based on the centroidal model can observe the planned foot placement positions, which allows it to anticipate the  terrain, without having direct access to the terrain map.

\vspace{-5pt}

\paragraph{Reinforcement Learning with Centroidal Model}
We apply RL on the centroidal model to synthesize the control policy. We use an actor-critic algorithm optimized with Proximal Policy Optimization~\cite{PPO}. 
The input to the policy is the observations $\{ s, \{r_i\}, \{\phi_i \mod T \} \}$,
where $s=\{p_z, \dot p, R, \omega\}$ is the state of the rigid body excluding displacement; $r_i = p_i - p$ is the stance foot position or swing-leg target position, relative to the center of mass; and $\phi_i \mod T $ is the normalized phase of foot $i$.
The reward is based on the rigid body state without concern for many of the issues that lead to more complex reward for RL solutions based on a full model.


Since we have the Euler integration result in analytic form, the centroidal dynamics can be implemented on a GPU to allow thousands of parallel simulations to collect data. A bottleneck for simulation with a control policy in our setting is that we need to solve a small QP for each time step. We currently find no off-the-shelf QP solver that can reliably solve thousands of small QPs in parallel. We therefore implement a custom parallel QP solver specific to our needs. The solver is implemented with PyTorch \cite{pytorch} for easy prototyping and is based on the interior point method \cite{convex-optimization}. 
In our experiments, we run $1600$ simulations in parallel, with each environment step taking around 0.25\,s on average, including reward computation and policy query. This allows for a data collection rate of 6-7k steps per second. The QP solves consume about 80\% of the time, thus in principle we could achieve a significant additional speedup with further improvements to the QP solver. 
The benefit becomes more obvious in the presence of irregular terrain such as a field of stepping stones.
Instead of simulating complex contact dynamics between a height field and the robot links, as in related work~\cite{2020-science-blindQuadruped, deepgait}, we need only select the appropriate foothold at negligible cost.

\vspace{-5pt}
\paragraph{Realization on the Full Model}
Since training the policy does not rely on any knowledge of how the low-level controller will be implemented on the full-order model, any controller can be used
as long as it can generate the joint commands to realize the required GRFs and the given swing foot placements.
Recent work uses another learned policy to track the behavior of some reduced-order models, e.g., \cite{SLIP-Cassie, deepgait}, but the policy will only be applicable to the specific robot it is trained on. We use a simple approach that only leverages knowledge of the kinematics. 

For stance foot control, we transform the GRFs into joint torques using the Jacobian transpose: $\tau_\text{stance} = J^T f$,
where $J$ is the stance feet position Jacobian with respect
to motor states (without considering the state of the floating base). 

We realize the foot placement strategy for the swing foot using a virtual force, $f_\text{PD}$, arising from a cartesian-space PD controller. 
If $(\phi_i \mod T) = 0$, we set the swing foot target to be the updated position $p_i$ as in the centroidal model; a spline curve parameterized by time is then constructed to connect the current foot position and $p_i$, with a given foot height clearance. 
The curve is tracked using  $\tau_\text{swing}=J^T f_\text{PD}$.
If early foot contact is detected, the torques for the leg are set to zero until the planned stance phase.
For the case of late contact, there is no special treatment, as a desired GRF will naturally result in leg extension.

\vspace{-5pt}
\paragraph{\algoName-full} During training, we can use the full model for simulation while keeping other components the same as \algoName. We call this variant: \algoName-Full. We use the same observation and action space and the action will be transformed into joint torques as discussed in the previous paragraph for simulation. There are trade off between using \algoName~and \algoName-full. \algoName~allows for easy parallel simulation at the cost of compromised model fidelity while \algoName-full provide precise modeling at the cost of more expensive data collection.
\section{Results}

\begin{figure*}[t]
    \centering
    \includegraphics[width=0.9\textwidth]{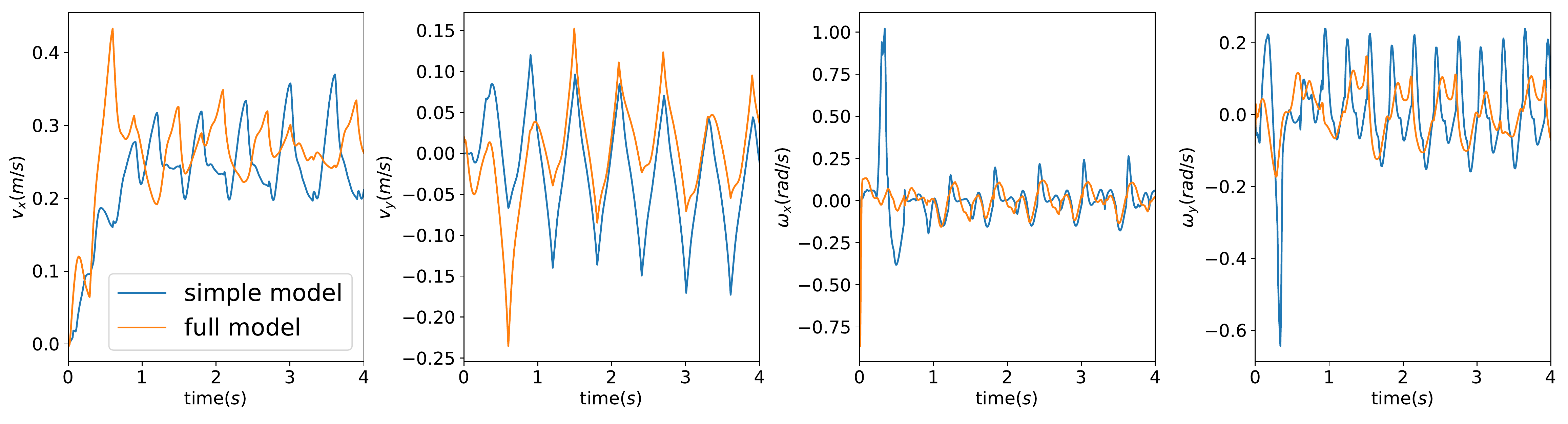}
    \caption{
    Linear velocity in the x-y plane as well as the roll and pitch velocity of full-order model and centroidal model of Laikago. The general behaviors are close.}
    \label{fig:full_vs_simple}
\end{figure*}


To demonstrate the generality of \algoName, we apply \algoName~to learn control policies for Laikago and A1.
The mass and inertia of the centroidal model as well as default foot placement locations and foot length constraints are adapted from the robot parameters provided by the manufacturer. Policies are parametrized with two-layer feedforward neural networks, where the hidden layer has size 128 with ReLU activation. TanH is applied to the output.

Our method benefits from an exceptionally simple reward structure. 
We use the following reward across all the tasks presented: $r = 0.5r_p + 0.5 r_o$,
where $r_p$ regularizes the linear velocity and body height, and $r_o$ regularizes the body orientation. We reward the policy for moving forward/backward in a straight line at desired velocity while keeping a default orientation. 
More specifically:
\begin{align*}
    r_p = &\exp(-10(\dot p-\dot p_{x,d})^2-50\dot p_y^2-50(p_z-p_{z,d})^2) \\
    r_o = & \exp(-10\norm{\Theta}^2),
\end{align*}
where $p_{z,d}$ is the nominal height of the robot, and $\Theta$ is the Euler angle representation of the robot orientation. 
Aside from locomotion on flat terrain, we also learn control policies for challenging tasks such as stepping stone traversal, navigating across a narrow balance beam, and two-legged balancing. We use IsaacGym~\cite{2020-IsaacGym} to test the policies on the simulated Laikago and A1 and we test some of the policies on a physical A1 robot to demonstrate robust sim-to-real. Please refer to the supplementary video for example motions.

\begin{figure*}[t]
    \centering
    \includegraphics[width=0.7\textwidth]{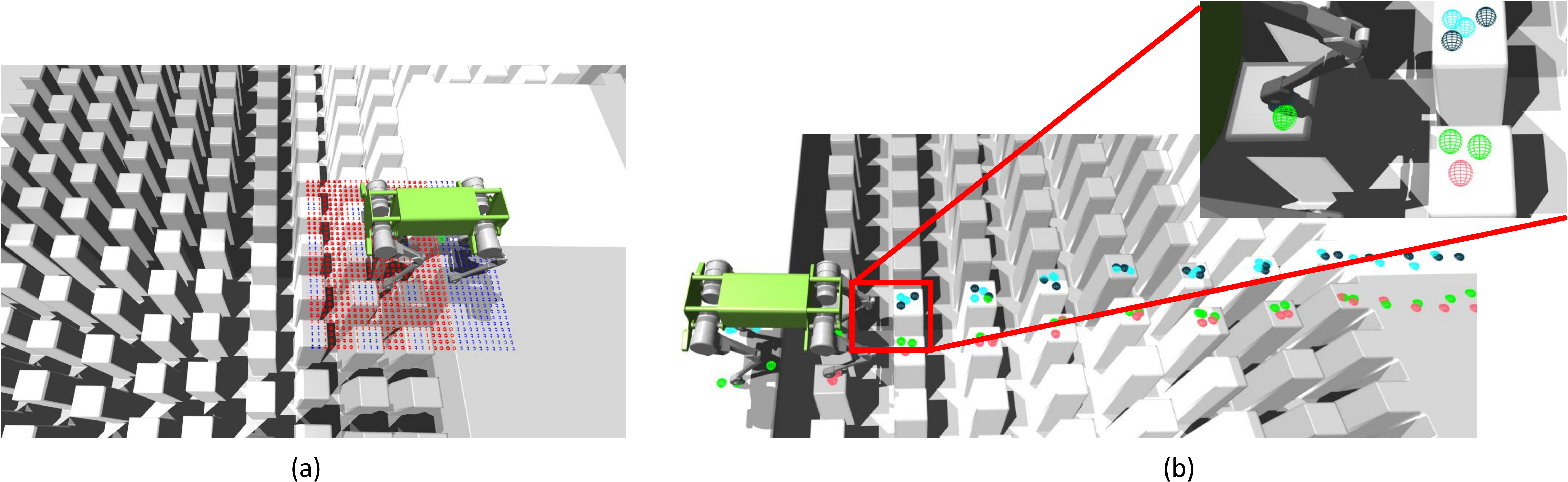}
    \caption{Illustration of the stepping stone. (a) The system gets a local map of the terrain around the robot, illustrated by the grid of red and blue points. Blue points indicate regions suitable for foot placement while the red points indicate otherwise. The edges of the stepping stones are also deemed unsuitable to encourage safety. The foot placement strategy chooses the feasible point that is closest to the default foot placement, as given by the Raibert heuristic. (b) Footstep pattern of Laikago trotting across a stepping stone terrain. Different colors indicate stepping locations of different feet.}
    \label{fig:stepping_stone}
\end{figure*}

\subsection{Learning Challenging Tasks with \algoName}

\paragraph{Flat Terrain}
We first demonstrate our system for the generation of locomotion on flat terrain. 
A trotting gait can be realized by synchronizing the leg phases of diagonal legs. We set the cycle time $T=60$ and swing time $T_\text{swing} = 30$. We initialize $\phi_1 = 0, \phi_4 = 0$ and $\phi_2 = 30, \phi_3 =30$ so the swing phase for nondiagonal legs are offset by half a cycle. Similarly, we can realize a walking gait with $T=120, T_\text{swing}=30$, and initialize $\phi_1=0,\phi_2=90,\phi_3=60,\phi_4=30$.

Control policies can be learned with the centroidal models of Laikago and A1 in 1--2 hours for trotting (up to $0.5$\,m/s)
and walking (up to $0.2$\,m/s). These policies are then deployed directly on their respective full-order robot models. 
We compare the behaviors of the simplified model and full model in Fig.~\ref{fig:full_vs_simple}. While the exact trajectories differ, they indicate similar general behavior.

Since we are learning with a centroidal model, the policy is agnostic to the morphology of the robot. We test the learned policy on a version of Laikago where the hind-leg knee configuration is inverted. The learned policy works directly for the inverted-configuration robot. We also train a trotting policy for the default Laikago, utilizing the full physics simulation and directly generate commands at the joint control level, similar to ~\cite{ 2020-RSS-laikago, 2021-icra-dynamics_randomization}. This policy with joint level action space works well with the default configuration of Laikago, as expected, but fails to generalize to the inverted configuration due to the learned control policy being very specific to the morphology it was trained on. 

We also test the generality of our approach by testing a policy trained with Laikago robot parameters on the A1 robot. 
The sizes of these two robots are not significantly different. We therefore choose not to apply scaling to the action space. However, we do scale the most relevant input features to the policy, namely the body height and foot positions.
This policy achieves a normalized reward of 0.93 on the Laikago robot and a normalized reward of 0.90 on the A1 robot. 
Directly applying the policy without scaling the input causes the A1 robot to fall over because the the input is not in the distribution seen during training. 
\vspace{-5pt}

\paragraph{Stepping Stones}
To showcase that \algoName~is compatible with tasks that require perception of the terrain, we train policies to drive the centroidal model across stepping stones. This terrain restricts the set of valid stepping locations, so the foot placement strategy needs to avoid infeasible locations on the ground. This task has been considered for quadrupedal locomotion in recent work 
using MPC
for trotting~\cite{control-barrier-stepping-stone} and static walking~\cite{ quadruped_planning}, 
and a simpler version of the task has been considered using RL~\cite{deepgait}. We demonstrate the generality of \algoName~by solving it with both Laikago and A1, with either a trotting or a walking gait. Fig.~\ref{fig:stepping_stone} illustrates the stepping stone problem. 
As before, we use the Raibert-style heuristic to plan foot placement.
However, if the planned foot placement is infeasible, the closest feasible position is chosen instead.

\begin{table}[t]
    \centering
\scriptsize
\begin{tabular}{l|c|c}
\toprule
\rowcolor[HTML]{CBCEFB}
\textbf{Method} & Energy & Motion quality \\
\midrule
\algoName(Flat) &$891 \pm 21$ & good \\
\rowcolor[HTML]{EFEFEF} 
\algoName-full (Flat) &$913 \pm 21$ & good \\
Centroidal PD (Flat)  &$869 \pm 4$ & good \\
\rowcolor[HTML]{EFEFEF} 
Joint RL (Flat) & $3476 \pm 75$ &bad\\
\hline
\algoName(Step) & $1080 \pm 46$ &good\\
\rowcolor[HTML]{EFEFEF} 
\algoName-full (Step) & $1113 \pm 52$ & good\\
Centroidal PD (Step) &$1460 \pm 70$ & bad \\
\bottomrule
\end{tabular}
\caption{Energy and motion quality of different methods. See the text for qualitative result for the motion quality assessment.}
\label{tab:baseline_comparison}
\end{table}

We train policies to traverse random stepping stone terrains with feasible footholds being placed between 10\,cm to 20\,cm apart for Laikago and 5\,cm to 15\,cm apart for A1. The training takes 4 hours, and the policies can be realized on the simulated full model of Laikago and A1.
The policies also generalize to stepping stone patterns not seen during training. We command the robot to walk across the stepping stones while turning, while the trained policy has only seen straight-line forward motion.
The policy successfully generalizes to this scenario. 

\vspace{-5pt}
\paragraph{Balance Beam}
Next we tackle the balance beam tasks to showcase the ability of our system to make control decisions that optimize performance over a time horizon. A similar capability was also demonstrated in~\cite{line_walking} using MPC method.

In this task, the robot must walk on a narrow path with limited foot placement choice in the lateral direction. We train a balance beam policy for the centroidal model of the A1. During training, we restrict the foot placement in world coordinates with $|p_{i, y}| \leq 0.05$\,m while the nominal leg distance between left and right foot is 0.2\,m.  
The $y$-coordinate of the rigid body $p_y$ is also provided to the policy to avoid drift. We train a trotting policy with desired velocity $\dot p_{x,d} = 0.1$\,m/s. The learned policy successfully transfers to the full simulated model of the A1 trotting across a balance beam.

\vspace{-5pt}
\paragraph{Two-Legged Balancing}
We then consider the two-legged balancing task. This is more challenging than the balancing beam task since the robot cannot regain balance by taking steps. 
In the two-legged balancing scenario, we set $\phi_1, \phi_4 = 31$ and $\phi_2, \phi_3 = 0$ for all time such that the left front leg and the right rear leg are set to be the stance legs. We also include $p_x, p_y$ as input to the policy. With the full-order model, the right front leg and the left hind leg are held in place in midair using position control, and the stance legs are controlled as before.


\subsection{Sim-to-Real Tests}
We transfer some of the policies to the physical A1 robot.
These policies do not rely on a full-body physics simulator and thus avoid the most common types of overfitting problems. Sim-to-real transfer is as straightforward as the reduced-order to full model transfer. This further demonstrates the generality and robustness of our approach. 


To generate gaits of different styles, we train a trotting policy on flat terrain with the default foot position $r_{\text{ref}, x} = \pm 0.1\,\mathrm{m}, r_{\text{ref}, y} = \pm 0.05$\,m, twice as close to the center of mass as the default. We set $T = 40$ and $T_\text{swing} = 20$ so the policy is trotting at a higher frequency. This policy can transfer to the full-order model in simulation and directly transfer to the physical A1. 
The learned policies are also robust to perturbations. 
In the video, we show the physical robot is able to handle obstacle-strewn terrain using a policy trained only on flat terrain, with normal foot spacing and stepping frequency.

While we train policies to traverse stepping stones, we are not yet able to realize this on hardware due to lack of a localization system. We design a proxy test where a gap of 12\,cm wide is placed in front of the robot.
The foot placement strategy is modified to generate footholds that avoid this gap. The stepping stone policy is able to generate commands that enable the stance legs to stabilize the robot while the swing legs take larger than usual steps.

\section{Evaluation}
\begin{figure}[t]
    \centering
    \includegraphics[width=0.7\textwidth]{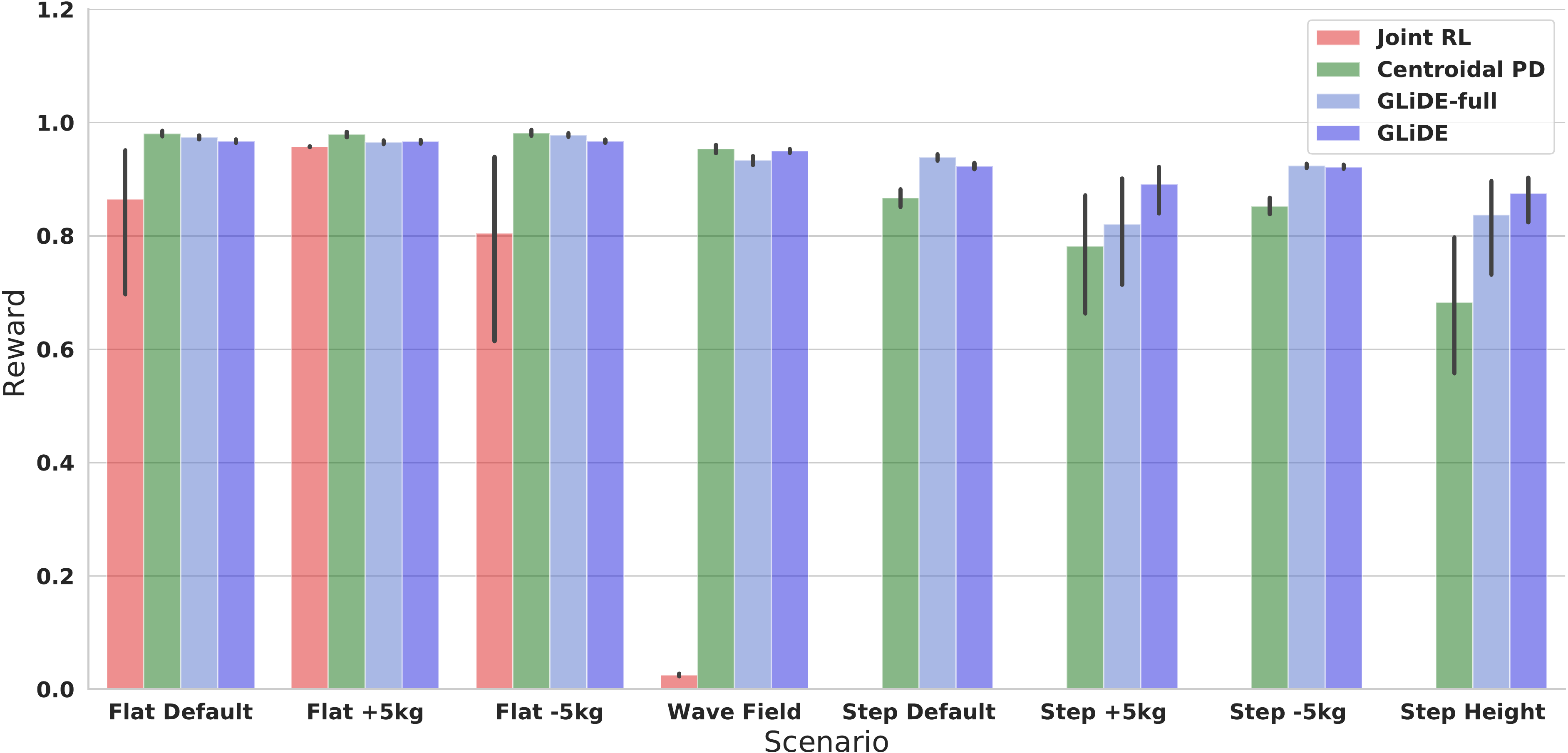}
    \caption{Normalized reward of different methods. \algoName~and \algoName-full are comparable to baseline in the default scenarios and perform better in challenging scenarios.}
    \label{fig:baseline_comparison}
\end{figure}

We compare \algoName~with alternatives using RL or MPC, as well as ablation on \algoName-full. Policies are trained with Laikago, on flat terrain (Flat) and stepping stone (Step). We test the performance of different policies under different test scenarios. The statistics of $10$ runs are recorded, where each run involves a policy running for 10 seconds in simulation. Variation between different runs include different initial body poses and different random seeds for terrain generation. Quantitative results are recorded in Fig.~\ref{fig:baseline_comparison} and Table~\ref{tab:baseline_comparison}.

\subsection{Test Scenarios}
\paragraph{Body Mass Perturbation} We perturb the mass of the body by $\pm 5$~kg, around 40\% of the default body mass and record the normalized reward of the policies.

\begin{figure*}[t]
    \centering
    \includegraphics[width=0.9\textwidth]{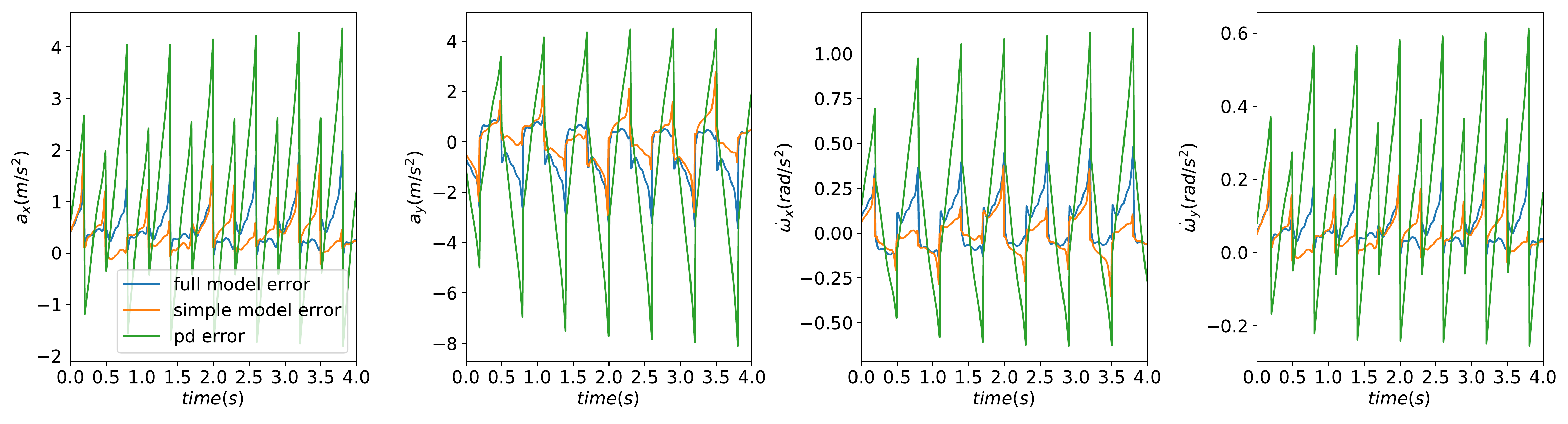}
    \caption{Comparison of difference between the desired and the actual body acceleration. Due to the underactuation of the systems, non-zero error is unavoidable. We plot the error of a \algoName~policy on the centroidal model and the full-order model, and plot the error of a centroidal PD controller on the full-order model. The \algoName~policy produces commands with a much smaller discrepancy compared to a centroidal PD controller.}
    \label{fig:acc_error}
\end{figure*}

\vspace{-5pt}
\paragraph{Height Variation} We record the normalized reward of the policies walking against height perturbation.
We create a wave field by adding sinusoidal height change along both x and y direction on the continous terrain. The wave period repeats in every 2.5~m and the peak to valley distance is 0.7~m
comparing to the robot normal height of 0.4~m. For stepping stone, we create a random height field by passing a white noise height change into a second order low-pass filter. This returns a medium level random 1D signal and we apply it along both x and y direction with different random seed. This creates slopes of up to 20 degrees.

\vspace{-5pt}

\paragraph{Energy Comparison} We record
$ \frac{1}{T}\sum_{t=0}^{T-1} \norm{\tau_t}^2$
where $\tau_t$ is the torque value of all joints at time step t, and $T$ is the total number of time steps in 10 seconds.

\vspace{-5pt}

\paragraph{Motion Quality} We record the the $z$ trajectory of the front left foot over 10 seconds and compute an energy spectral density in Fourier space.

\subsection{Evaluation Results}

\paragraph{RL baseline: Joint RL} We train policies on the full model with joint torque as action space on flat terrain using the same reward. While these policies are able to obtain similar performance in term of reward, the policies fail to generalize to other scenarios
and favor high-frequency shuffle gaits. We record the $z$ trajectory of the front left foot for 10~$s$ and compute an energy spectral density in Fourier space. The fraction of energy for $f>10$~Hz is $0.13$ for \algoName~and $0.68$ for joint RL. Reward tuning is necessary to obtain suitable behavior, e.g, \cite{2019-science-sim2realAnyaml}. 

\vspace{-5pt}

\paragraph{MPC baseline: Centroidal PD}
As an alternative to \algoName, we implement a Centroidal PD controller used in \cite{cheetah3_design_control, contact-adaptive, StarlEth}; with $a_d$ is computed as

\begin{equation}
a_d = \left[\begin{array}{c}\ddot p_d \\ \dot \omega_d \end{array} \right] = k_p \left[\begin{array}{c}p_d-p \\ \Theta_d-\Theta \end{array} \right]+k_d \left[\begin{array}{c}\dot p_d-\dot p \\ \omega_d-\omega \end{array} \right],
\end{equation}
with $p_d=[0, 0, p_{z,d}], \dot{p}_d = [\dot p_{x,d}, 0, 0]$, $\Theta_d = [0, 0, 0], \omega_d = [0, 0, 0]$ being the desired behavior while $k_p$ and $k_d$ are diagonal matrices defining the PD gain.
Both \algoName~and Centroidal PD can produce the trotting and walking gaits on flat ground. However, Centroidal PD produces more aggressive motion. In Fig.~\ref{fig:acc_error}, we plot $Mf - a$, the error term for the QP problem (Equation \ref{eq:QP}), for both the learned policy and the cenotridal PD controller with a trotting gait. Due to underactuation, the error is unavoidable, but the learned policy incurs much smaller error compared to a centroidal PD controller.

For the stepping stone tasks, centroidal PD can be tuned to navigate across the stepping stone terrain with $T=40, T_\text{swing}=20$ with a trotting gait on Laikago, which is unrealistically fast for the hardware. We failed to find a Centroidal PD controller that achieves a more realistic stepping frequency. 
We believe the failure is caused by aggressive tracking of the command, whereas the learned policy prioritizes safety over tracking error. 
For both the balance beam task and the two-legged balancing task, centroidal PD falls sideways due to underactuation. Specialized methods are necessary to solve different tasks using MPC, e.g., \cite{line_walking, control-barrier-stepping-stone}, while \algoName~is able to solve all the tasks presented.

\vspace{-5pt}

\paragraph{\algoName-full} 
For ablation on \algoName, we also use \algoName-full to train policies for flat terrain and stepping stone. The performance of \algoName-full policies in principle will be an upper bound for the performance of \algoName~policies in the default scenario since the training environment and the testing environment is identical for \algoName-full. In Figure~\ref{fig:baseline_comparison}, we show that \algoName~and \algoName-full demonstrate similar performance on most testing scenarios while \algoName~performs more consistently on challenging scenarios such as Step Height and Step $+5$~kg.

\section{Discussion and Future Work}
We present a framework that combines RL with a centroidal model to learn policies for quadrupedal locomotion, as a simple (and therefore highly practical) alternative to the common strategy of using RL with a high-fidelity model. We validate our approach by efficiently learning locomotion policies for different quadrupedal robots and solving challenging tasks such as locomotion across stepping stones, two-legged balancing, and locomotion across a balance beam.  We further demonstrate successful sim-to-real transfer. The method contributes to the understanding of what is required, in terms of model-fidelity, for RL-based skill design for quadrupedal robots.


There remain constraints that are not captured by a simplified model, including joint limits and torque limits. 
These could be taken into account by using the full model for training,
with the caveat that the policy becomes specific to the full model.
Obtaining maximal performance may also require explicitly taking these limits into account, as can in 
principal be achieved using full-model RL.



We rely on a simple heuristic for foot placement. This limits our ability to deal with more challenging terrain, e.g., more challenging variations of the stepping stone task. Incorporating systems~\cite{DeepQ-Stepper, RLOC} that directly learn foot placement will be crucial for further improving the capability of our system. Our system can in principle also handle terrains with height variations, e.g., for climbing stairs. We are in the process of extending our framework for learning to traverse more general terrains, as well as incorporating a vision system onto the physical A1 robot to enable sim-to-real transfer for tasks such as stepping stone traversal.

Currently we need to train different policies for different scenarios. A unified policy that can handle all scenarios will be crucial for real world adoption. This can potentially be achieved via progressive learning~\cite{progressive_rl}.



%
%
%
%

\end{document}